\pdfoutput=1
\documentclass[11pt]{article}

\usepackage[]{EMNLP2023}

\usepackage{times}
\usepackage{latexsym}

\usepackage[T1]{fontenc}
\usepackage[utf8]{inputenc}

\usepackage{microtype}

\usepackage{inconsolata}

\usepackage{graphicx}
\usepackage{amsmath}
\usepackage{multirow}
\usepackage{subcaption}

\newcommand{\az}[1]{\color{black} #1}

\def\tokgen{\mathcal{T}_{gen}}
\def\tok100{\mathcal{T}_{100}}

\title{Multi-word Tokenization for Sequence Compression}
\author{Leonidas Gee\\
  University of Sussex, United Kingdom \\ \texttt{jg717@sussex.ac.uk}
  \And
  Leonardo Rigutini\\
  expert.ai, Siena, Italy\\
  \texttt{lrigutini@expert.ai}\\
  \AND
  Marco Ernandes\\
  expert.ai, Siena, Italy \\ \texttt{mernandes@expert.ai}
  \And
  Andrea Zugarini\\
  expert.ai, Siena, Italy \\ \texttt{azugarini@expert.ai}
}

\begin{document}

\maketitle

%%% Abstract %%%

\begin{abstract}
Large Language Models have proven highly successful at modelling a variety of tasks. However, this comes at a steep computational cost that hinders wider industrial uptake. {\az In this paper, we present MWT: a Multi-Word Tokenizer that goes beyond word boundaries by representing frequent multi-word expressions as single tokens. MWTs produce a more compact and efficient tokenization that yields two benefits:} (1) Increase in performance due to a greater coverage of input data given a fixed sequence length budget; (2) Faster and lighter inference due to the ability to reduce the sequence length with negligible drops in performance. Our results show that MWT is more robust across shorter sequence lengths, thus allowing for major speedups via early sequence truncation.
% Large Language Models (LLMs) have proven highly successful at modelling a variety of tasks. However, this comes at a steep computational cost that hinders wider industrial uptake. In this paper, we propose a method for compressing the input sequence of a LLM using Multi-word Expressions (MWEs). This produces two benefits: (1) Increase in performance due to a greater coverage of input data given a fixed sequence length budget; (2) Faster and lighter inference due to the ability to reduce the sequence length with negligible drops in performance. Our results show that tokenization using MWEs is more robust across shorter sequence lengths, thus allowing for major speedups via early sequence truncation.
\end{abstract}

%%% Introduction %%%

\section{Introduction}\label{section:introduction}
The field of Natural Language Processing (NLP) has seen major breakthroughs with the advent of Large Language Models (LLMs)~\citep{Transformer, BERT, touvron2023llama,openai2023gpt4}. Despite their successes, LLMs like ChatGPT~\citep{openai2023gpt4, GPT3} possess hundreds of billions of parameters that entail enormous computational cost by design. {\az Traditional model compression methods such as Knowledge Distillation~\citep{Distillation}, Pruning~\citep{16Heads,Prune}, and Quantization~\citep{QBERT, Precision} have focused on creating lighter models either by shrinking the architectural size or by reducing the number of FLOPs.}
% Maintaining and deploying such models becomes rapidly unsustainable when hundreds of different instances are required. As such, many small to medium sized enterprises struggle to adopt LLMs for their business needs.

{\az Recently, LLMs have been shown to produce impressive performance on inputs that have been carefully designed to contain all the necessary information for a given instruction. As such, there is an increasing trend in designing longer and longer prompts that has led to a significant rise in computational cost. To address this, interest has grown in compressing the input sequences from the tokenizer~\citep{FVT, mu2023learning, petrov2023language}. Indeed, various works have shown the importance of tokenization in determining the length of a sequence in specialized domains~\citep{FVT} or on underrepresented languages~\citep{petrov2023language}.}

\begin{figure*}
    \fbox{
        \begin{minipage}{2.01\columnwidth}
        \small
        \textbf{Input:}
        \texttt{an energizable member is operably coupled to the outer sleeve .}
        \\ 
        \\
        \textbf{$\tokgen$:}
        \texttt{an, en, \#\#er, \#\#gi, \#\#zable, member, is, opera, \#\#bly, coupled, to, the, outer, sleeve, .}
        \\
        \\
        \textbf{$\tokgen^{1000}$:}
        \texttt{an, en, \#\#er, \#\#gi, \#\#zable, \textbf{\color{blue} member\_is}, opera, \#\#bly, \textbf{\color{blue} coupled\_to}, \textbf{\color{blue} the\_outer}, sleeve, .}
        \\
        \\
        \textbf{$\tok100$:}
        \texttt{an, \textbf{\color{orange}energizable}, member, is, \textbf{\color{orange} operably}, coupled, to, the, outer, sleeve, .}
        \\
        \\
        \textbf{$\tok100^{1000}$:}
        \texttt{an, \textbf{\color{orange} energizable}, \textbf{\color{blue} member\_is}, \textbf{\color{orange} operably}, \textbf{\color{blue} coupled\_to}, \textbf{\color{blue} the\_outer}, sleeve, .}
        \end{minipage}
    }
    \caption{Tokenization using generic $\tokgen$ and adapted $\tok100$ tokenizers. $\tokgen^{1000}$ and $\tok100^{1000}$ are extended with the top-1000 bigrams. Tokens obtained with domain-adaptation or MWT are highlighted in orange and blue respectively. MWTs are shown to be highly complementary to existing tokenizers for sequence compression.}
    \label{figure:tokenization}
\end{figure*}

% To address this, significant effort has been made in compressing the size of LLMs, while retaining as much of their initial performance as possible. Several successful methods have been developed in this direction such as Knowledge Distillation~\citep{Distillation}, Pruning~\citep{16Heads,Prune}, Quantization~\citep{QBERT, Precision}, and most recently, Vocabulary Transfer~\citep{FVT}.

In this paper, we propose a method for reducing the computational cost of a LLM by compressing the textual inputs using Multi-Word Tokenizers (MWTs). {\az To achieves this, we enrich the vocabulary of the tokenizer with statistically determined multi-word expressions. By encoding the frequent n-grams with single tokens, the sequences produced are both shorter and more informative}, thus allowing for major speedups via early sequence truncation. Additionally, MWTs are shown to be compatible with the aforementioned traditional compression methods. Experimentally, {\az we assess MWTs on three text classification datasets. We show how our approach still performs well when combined with distilled models~\citep{DistilBERT} and other sequence compression techniques~\citep{FVT}.} The code for our paper is publicly available\footnote{\url{https://github.com/LeonidasY/fast-vocabulary-transfer/tree/emnlp2023}}.

The rest of the paper is organized as follows. First, we review the related works in Section~\ref{section:relworks}. Then, we describe our approach in Section~\ref{section:methodology} and present the experiments in Section~\ref{section:experiments}. Finally, we draw our conclusions in Section~\ref{section:conclusion}.

%%% Related Works %%%

\section{Related Works}\label{section:relworks}
% {\az An enormous effort has been dedicated in making LLMs more efficient and less resource demanding. Compression techniques are evolving rapidly.}
Most model compression research falls into one of the following categories: Knowledge Distillation~\citep{Distillation, DistilBERT, TinyBERT, MiniLM, MobileBERT}, Pruning~\citep{Prune, 16Heads}, and Quantization~\citep{QBERT}. The family of approaches is somewhat complementary and can be applied individually or jointly. Each approach alters the model's size to obtain a more efficient architecture. {\az Differently, other works such as FlashAttention~\citep{dao2022flashattention} seek to optimize a model's implementation. In particular, LLMs are sped up by reducing the number of memory accesses for the self-attention mechanism.}

{\az 
\paragraph{Sequence Compression.}
An emerging direction for reducing the cost of LLMs involves the designing of shorter input sequences. Prompting techniques such as \citet{mu2023learning} compress repetitive lengthy prompts into gist tokens. Other works emphasize the role of tokenization in sequence compression. In \citet{petrov2023language}, the authors show how the tokenizer of most LLMs strongly favor the English language over other languages. For underrepresented languages, the same translated sentence may consist of inputs that are up to 15 times longer. Analogously, \citet{FVT} investigated the tokenization efficiency of general-purpose tokenizers in vertical domains such as medicine and law. They proposed a transfer learning technique that adapts the vocabulary of a LLM to specific language domains. An effect of a dedicated vocabulary is a more efficient tokenization that reduces the number of sub-word tokens in a sequence.
}

In this work, we push this effect further, going beyond word boundaries by introducing Multi-Word Expressions (MWEs) in the form of n-grams into the tokenizer as shown in Figure~\ref{figure:tokenization}. The underlying intuition behind this is that a more compact tokenization can save computations by allowing the model to process shorter sequences without a significant loss of information. The usage of MWEs is not novel with several works~\citep{Phrase, MWE, MBPE} introducing phrases or n-grams to improve the quality of machine translation. In~\citet{MBPE}, the authors generalized BPE~\citep{BPE} to multi-word tokens. However, to the best of our knowledge, we are the first to investigate MWEs in the context of sequence compression.

%%% Methodology %%%

\section{Multi-word Tokenizer}\label{section:methodology}
\begin{figure*}[ht]
    \centering
    \includegraphics[scale=0.85]{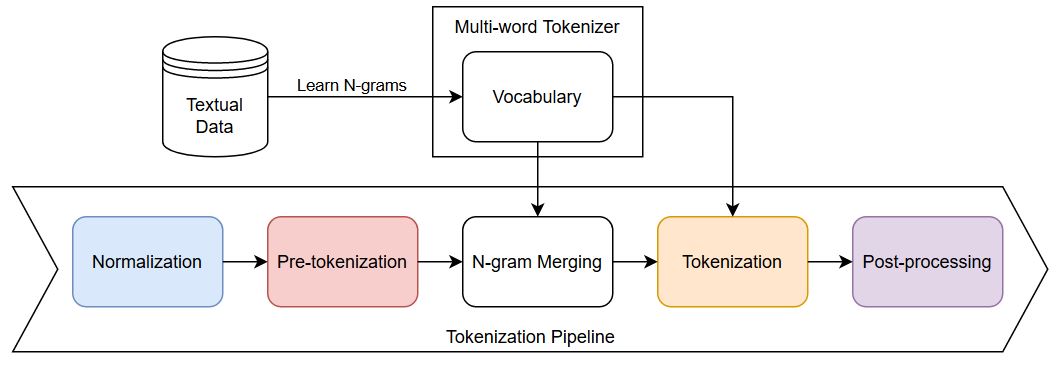}
    \caption{Sketch of the Multi-word Tokenizer pipeline. First, n-grams are statistically learned from the training set. Then, the top-K n-grams are added to the vocabulary of the tokenizer. N-grams are merged from left to right within a sequence after pre-tokenization.}
    \label{figure:mwt_sketch}
\end{figure*}

Tokenization is a necessary step in the feeding of textual data to a LLM. Typically, tokenizers split a text into a sequence of symbols which can be entire words or only subparts. To do this, a vocabulary is first constructed by statistically learning the most frequent tokens from a large general-purpose corpus~\citep{BPE, WordPiece, SentencePiece}. The resulting tokenizer can then be used to segment an input text by greedily looking for the solution with the least number of tokens. 
Building upon this, we inject into the tokenizer new symbols formed by n-grams of words. We do this by first selecting the most frequent n-grams to include in its vocabulary. Then, we place an n-gram merging step within the tokenization pipeline as sketched in Figure~\ref{figure:mwt_sketch}. The added n-grams will be treated as single tokens further down the tokenization pipeline.

\paragraph{N-gram Selection.}
In order to maximize the sequence reduction, we statistically estimate the top-K most frequent n-grams in a reference training corpus. Although the approach is greedy, hence sub-optimal, it still effectively yields significant compression while being extremely fast and easy to compute. More formally, given a corpus $\mathcal{D}$ and $N \geq 2$, we compute all the possible n-grams $g_n \in \mathcal{D}$, where $n=2,\ldots, N$. Then, we count their frequency $f(g_n), \forall g_n \in \mathcal{D}$. The $K$ most frequent n-grams $\mathcal{G}_K$ are included in the vocabulary $\mathcal{V} \leftarrow \mathcal{V} \cup \mathcal{G}_K$ of the tokenizer $\mathcal{T}$.

\paragraph{Fast Vocabulary Transfer.} Given that the vocabulary of the tokenizer has changed, the newly added symbols $\mathcal{G}_K$ must be included into the embedding matrix of the language model as well. To avoid retraining the entire model from scratch which is highly resource-demanding, or a random initialization of new tokens which would perform poorly, we make use of Fast Vocabulary Transfer (FVT) instead~\citep{FVT}. 

\begin{figure}
    \centering
    \includegraphics[scale=0.7]{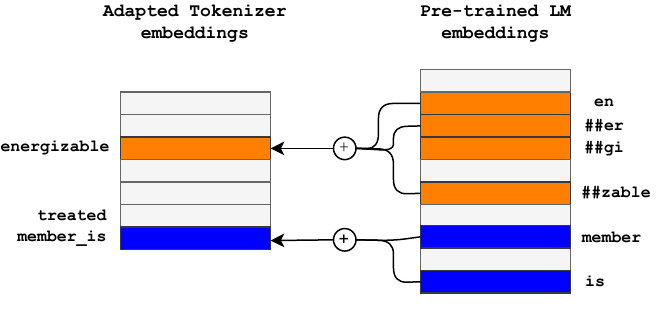}
    \caption{Fast Vocabulary Transfer. The pre-trained embeddings of existing tokens are combined to form the embeddings of the newly adapted vocabulary.}
    \label{figure:fvt_sketch}
\end{figure}

FVT is a transfer learning technique that assigns embeddings to new tokens by combining existing elements of the embedding matrix as shown in Figure~\ref{figure:fvt_sketch}. After initializing the multi-word embeddings with FVT, we found it beneficial to tune the model with Masked-Language Modeling (MLM) as done by~\citet{FVT}. We believe this is helpful as it aids the model in further readjusting the embeddings of the new tokens.

\renewcommand{\arraystretch}{1.3}
\begin{table*}[ht]
    \centering
    \begin{tabular}{ccccccccc}
        \hline
        \textbf{Dataset}    & $\tokgen$     & $\tokgen^{1000}$  & $\tokgen^{2500}$  & $\tokgen^{5000}$  & $\tok100$     & $\tok100^{1000}$  & $\tok100^{2500}$  & $\tok100^{5000}$  \\
        \hline
        \textbf{ADE}    & 31    & 26    & 25    & 23    & 21    & 18    & 17    & 16    \\
        \textbf{LEDGAR} & 155   & 118   & 107   & 98    & 131   & 97    & 90    & 84    \\
        \textbf{PATENT} & 134   & 110   & 105   & 100   & 118   & 94    & 90    & 86    \\ 
        \hline
    \end{tabular}
    \caption{Average sequence length from tokenization. The generic $\tokgen$ and adapted $\tok100$ tokenizers are extended with varying top-Ks of 1000, 2500, and 5000.}
    \label{table:seq_len}
\end{table*}

%%% Experiments %%%

\section{Experiments}\label{section:experiments}
Given a fixed number of tokens, a more compact input sequence preserves a greater amount of information. This can be used to either achieve a better performance with limited benefits in speedup, or vice versa, i.e. making the model faster with negligible drops in performance. The experiments aim to analyze how these two aspects interact with one another. {\az We focus on text classification as it is a problem of particular interest for many industry-oriented applications.}

\subsection{Experimental Setup}
Our experiments were conducted on the cased versions of $\text{BERT}_{base}$~\citep{BERT} and $\text{DistilBERT}_{base}$~\citep{DistilBERT}. Additionally, we consider an adapted tokenizer with a vocabulary size equal to that of the generic tokenizer from a pre-trained model as done by~\citet{FVT}. We refer to the generic and adapted tokenizers as $\tokgen$ and $\tok100$ respectively. Both tokenizers are extended with the top-K n-grams of 1000, 2500, and 5000. Overall, we compare eight different tokenizers indicated as: $\tokgen, \tokgen^{1000}, \tokgen^{2500}, \tokgen^{5000}$ and $\tok100, \tok100^{1000}, \tok100^{2500}, \tok100^{5000}$.

\paragraph{Implementation Details.}
We train each model with 5 different random initializations. The macro-F1 and inference speedup are measured as metrics. The average of all 5 initializations is taken as the final value of each metric. The inference speedup measurements were done on a \mbox{V100-PCIE} GPU with 16GBs of dedicated RAM.

Following~\citet{FVT}, we first apply one epoch of MLM using the in-domain dataset. Next, the model is fine-tuned for 10 epochs with early stopping on the downstream task. We set the initial learning rate to $3\cdot10^{-5}$ for both MLM and downstream fine-tuning, while the batch size is set to 8 and 32 for MLM and downstream fine-tuning respectively.

\paragraph{Choice of N.}
An important hyperparameter is N, i.e. the maximum number of words constituting an n-gram. In our experiments, N is set to 2 as we believe that using bigrams only provides better generalization properties. Increasing the value of N may lead to an overspecialization of n-grams which could overfit on small textual corpora.

\begin{figure*}[ht]
    \centering
    \includegraphics[scale=0.13]{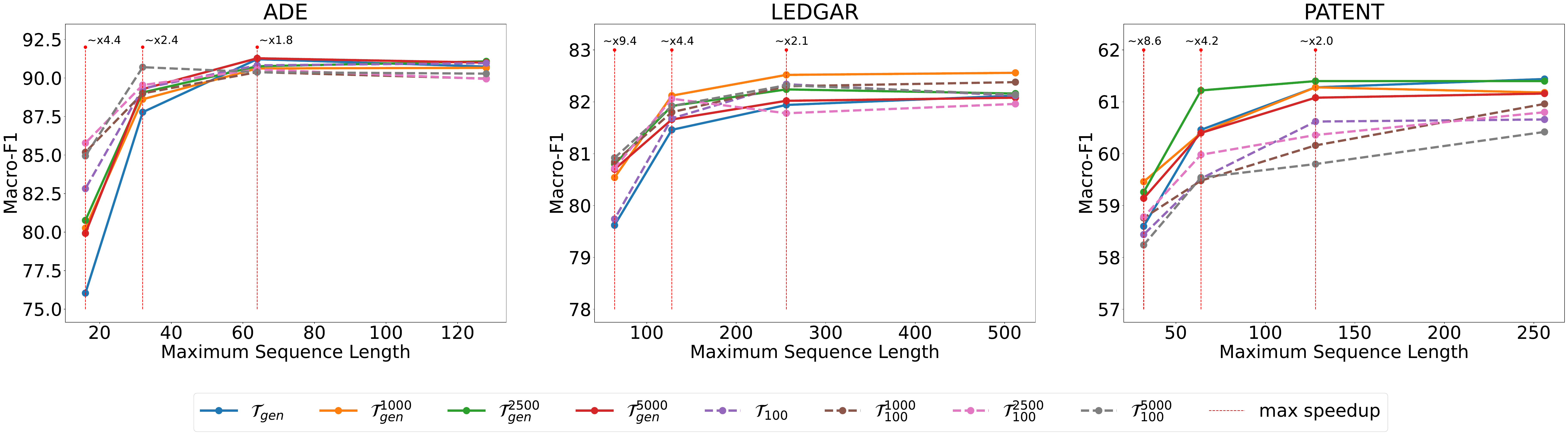}
    \caption{Plot of macro-F1 against maximum sequence length. The generic $\tokgen$ and adapted $\tok100$ tokenizers are represented by solid and dashed lines respectively. MWTs are shown to be more robust on shorter sequence lengths, thus allowing for major speedups via early sequence truncation.} 
    \label{figure:results}
\end{figure*}

\subsection{Datasets}
To determine the effectiveness of MWTs, we select 3 different text classification tasks from diverse linguistic domains, namely medical (ADE), legal (LEDGAR), and tech (PATENT).

\paragraph{ADE.} A sentence classification dataset of determining whether a sentence is Adverse Drug Event (ADE)-related or not~\citep{ADE}. The sentences are characterized by the presence of medical terminologies of drugs and their adverse effects. We use the same train, validation, and test splits as in~\citet{FVT}.

\paragraph{LEDGAR.} A document classification dataset of contracts obtained from the US Securities and Exchange Commission (SEC) filings~\citep{LEDGAR}. The task is to determine whether the main topic of the contract provision from a set of 100 mutually-exclusive labels. The dataset is also part of LexGLUE~\citep{LexGLUE}, which is a benchmark for legal language understanding.

\paragraph{PATENT.} A document classification dataset\footnote{\url{https://huggingface.co/datasets/ccdv/patent-classification}} of US patent applications filed under the Cooperative Patent Classification (CPC) code~\citep{PATENT}. A human written abstractive summary is provided for each patent application. The task is to determine the category that a patent application belongs to from 9 unbalanced classes.

\subsection{Results}

\paragraph{Preliminary Analysis.}
Before measuring the effects of MWTs on LLMs, we analyze how the average sequence length changes for each dataset depending on the tokenizer. From Table~\ref{table:seq_len}, increasing the top-K most frequent n-grams naturally yields a greater compression. However, even a 1000 bigrams is enough to achieve a reduction of about 20\%. When multi-words are combined with an adapted tokenizer $\tok100$, the joint sequence narrowing effects appear to be highly complementary, achieving a compression rate close to 50\% in ADE. In practice, a 50\% reduction means that on average we can store the same amount of text in half the sequence length. Consequently, we could in principle reduce a LLM's maximum sequence length by a factor of 2.

\renewcommand{\arraystretch}{1.3}
\begin{table*}[ht]
    \centering
    \begin{tabular}{ccccccc}
        \hline
        \multirow{2}{*}{\textbf{Method}} & 
        \multicolumn{2}{c}{\textbf{ADE}} & \multicolumn{2}{c}{\textbf{LEDGAR}} & \multicolumn{2}{c}{\textbf{PATENT}} \\
        
        &   $\Delta$\textbf{F1} &   \textbf{Speedup}
        &   $\Delta$\textbf{F1} &   \textbf{Speedup} 
        &   $\Delta$\textbf{F1} &   \textbf{Speedup}    \\       
        \hline
        
        $\tokgen$           & 90.74 $\pm$ 0.84          & 1.00          & 82.12 $\pm$ 0.33          & 1.00          & \textbf{61.44 $\pm$ 0.38} & 1.00          \\       
        \hline

        $\tokgen^{1000}$    & -0.09 $\pm$ 0.70          & 1.32          & \textbf{0.54 $\pm$ 0.24}  & 1.14          & -0.42 $\pm$ 0.54          & 1.11          \\
        $\tokgen^{2500}$    & \textbf{0.37 $\pm$ 0.54}  & 1.38          & 0.05 $\pm$ 0.44           & 1.23          & -0.07 $\pm$ 0.46          & 1.16          \\
        $\tokgen^{5000}$    & 0.29 $\pm$ 0.68           & 1.43          & -0.05 $\pm$ 0.41          & 1.33          & -0.46 $\pm$ 0.69          & 1.19          \\    
        \hline

        $\tok100$           & 0.24 $\pm$ 0.67           & 1.51          & 0.00 $\pm$ 0.41           & 1.10          & -1.27 $\pm$ 0.39          & 1.06          \\
        $\tok100^{1000}$    & -0.86 $\pm$ 1.21          & 1.71          & 0.32 $\pm$ 0.58           & 1.36          & -0.78 $\pm$ 0.62          & 1.24          \\
        $\tok100^{2500}$    & -0.88 $\pm$ 0.72          & 1.78          & -0.19 $\pm$ 0.57          & 1.47          & -1.04 $\pm$ 0.42          & 1.30          \\
        $\tok100^{5000}$    & -0.51 $\pm$ 0.65          & \textbf{1.79} & 0.02 $\pm$ 0.58           & \textbf{1.57} & -1.66 $\pm$ 0.44          & \textbf{1.34} \\      
        \hline
    \end{tabular}
\caption{Absolute values of BERT fine-tuned on the downstream task using a sequence length of 128, 512 and 256 for ADE, LEDGAR and PATENT respectively. $\tokgen$ is shown on the first row, while relative values to $\tokgen$ are shown on subsequent rows.}
\label{table:relative}
\end{table*}

\paragraph{Multi-word Tokenization.}
As a first evaluation, we assess the macro-F1 and inference speedups achieved by fine-tuned BERT models with multi-word tokenizers: $\tokgen^{1000}, \tokgen^{2500}, \tokgen^{5000}$. The pre-trained BERT with a generic tokenizer $\tokgen$ is considered as the reference model. From Table~\ref{table:relative}, MWTs are shown to either improve the reference performance or induce a relatively negligible degradation. At the same time, the sequence compression from MWTs yields a natural speedup that depending on the dataset varies from about x1.1 to x1.4.

\paragraph{MWT and Domain Adaptation.}
Additionally, we investigate the application of MWTs with tokenizers adapted to the dataset: $\tok100^{1000}, \tok100^{2500}, \tok100^{5000}$. With the exception of PATENT, most models are shown to achieve significant inference speedups of up to x1.8 with minimal degradation in performance from Table~\ref{table:relative}. We hypothesize that this is due to the fact that the language domain of PATENT is not as specialized as ADE and LEDGAR, which reduces the benefits of using an adapted tokenizer.

\paragraph{MWT and Truncation.} 
Based on the preliminary analysis, we analyze how truncating sequences with different maximum lengths affects both the performance and inference speedup. Reducing the maximum sequence length has a double impact on the inference speedup given a fixed amount of resources. First, latency linearly grows with respect to the sequence length. Second, reducing the sequence length releases GPU resources that can be used to enlarge the batch size. We consider 4 maximum sequence lengths for each dataset by progressively halving the initial maximum sequence length, i.e. $\{128,64,32,16\}$ for ADE, $\{256,128,64,32\}$ for LEDGAR, and $\{512,256,128,64\}$ for PATENT.

From Figure~\ref{figure:results}, we can see the performance of $\tokgen$ dropping more rapidly than MWTs as truncation increases (maximum sequence length decreases). In the extreme 8-times truncation, the performance of $\tokgen$ falls dramatically for both ADE and LEDGAR. However, MWTs are shown to be more robust to truncation, hence their degradation in performance is smoother and without sudden collapses. In both ADE and LEDGAR, a 4-times truncation leads to nearly identical or better performance, while bringing significant inference speedups of $\sim$x2.4 and $\sim$x4.4 respectively. If a certain performance degradation is acceptable, the inference speedup can be maximized, reaching up to $\sim$x9.4 in LEDGAR.

\paragraph{MWT and Distillation.} 
Additionally, we investigate the interaction between sequence compression and knowledge distillation in Table~\ref{table:distillation}. To this end, we utilize a DistilBERT model with MWTs. For simplicity, we restrict our analysis to LEDGAR and to a single multi-word tokenizer $\tokgen^{2500}$ on different maximum sequence lengths. From the table, our MWT is shown to retain most of its performance with a quarter of the sequence length and an inference speedup of $\sim$x8.8. Even with an extreme sequence truncation to only 64 tokens, we can still achieve a $\sim$x18.1 inference speedup with only a 2.7\% drop in relative performance. 

\renewcommand{\arraystretch}{1.3}
\begin{table}[ht]
    \centering
    \begin{tabular}{cccc}
        \hline      
        \textbf{Model}                  & \textbf{Length}   & $\Delta$\textbf{F1}   & \textbf{Speedup}  \\
        \hline
    
        $\tokgen$                       & 512               & 82.12                 &   1.00            \\ 
        
        \hline
        
        Distil. + $\tokgen$             & 512               & -0.78                 &   2.43            \\
        Distil. + $\tokgen^{2500}$      & 128               & -0.32                 &   8.81            \\
        Distil. + $\tokgen^{2500}$      & 64                & -2.70                 &   18.13           \\
    
        \hline            
    \end{tabular} 
    \caption{The macro-F1 and inference speedup results on LEDGAR with DistilBERT. MWTs are shown to be highly compatible with distilled models.}
    \label{table:distillation}
\end{table}

%%% Conclusion %%%

\section{Conclusion}\label{section:conclusion}
In this work, we proposed a sequence compression approach that reduces textual inputs by exploiting the use of multi-word expressions drawn from the training set according to their top-K frequencies. We conducted an investigation on 3 different datasets by evaluating each model in conjunction with other compression methods~\citep{FVT, DistilBERT}. Our approach is shown to be highly robust to shorter sequence lengths, thus yielding a more than x4 reduction in computational cost with negligible drops in performance. In the future, we expect to extend our analysis to other language models and tasks such as language generation in the scope of sequence compression.

\section{Limitations}
As demonstrated in the paper, MWTs work well on text classification problems. Despite not having conducted experiments on generative tasks, there are no limitations in extending MWTs to them. Differently, the application of MWTs to token classification problems can be challenging. Specifically, when merging multiple words together, it is unclear how to label such fused tokens.

\section*{Acknowledgements}
This work was supported by the IBRIDAI project, a project financed by the Regional Operational Program “FESR 2014-2020” of Emilia Romagna (Italy), resolution of the Regional Council n. 863/2021.
% This document has been adapted by Yue Zhang, Ryan Cotterell and Lea Frermann from the style files used for earlier ACL and NAACL proceedings, including those for 
% ACL 2020 by Steven Bethard, Ryan Cotterell and Rui Yan,
% ACL 2019 by Douwe Kiela and Ivan Vuli\'{c},
% NAACL 2019 by Stephanie Lukin and Alla Roskovskaya, 
% ACL 2018 by Shay Cohen, Kevin Gimpel, and Wei Lu, 
% NAACL 2018 by Margaret Mitchell and Stephanie Lukin,
% Bib\TeX{} suggestions for (NA)ACL 2017/2018 from Jason Eisner,
% ACL 2017 by Dan Gildea and Min-Yen Kan, NAACL 2017 by Margaret Mitchell, 
% ACL 2012 by Maggie Li and Michael White, 
% ACL 2010 by Jing-Shin Chang and Philipp Koehn, 
% ACL 2008 by Johanna D. Moore, Simone Teufel, James Allan, and Sadaoki Furui, 
% ACL 2005 by Hwee Tou Ng and Kemal Oflazer, 
% ACL 2002 by Eugene Charniak and Dekang Lin, 
% and earlier ACL and EACL formats written by several people, including
% John Chen, Henry S. Thompson and Donald Walker.
% Additional elements were taken from the formatting instructions of the \emph{International Joint Conference on Artificial Intelligence} and the \emph{Conference on Computer Vision and Pattern Recognition}.

\bibliography{emnlp2023}
\bibliographystyle{acl_natbib}

\appendix
\section{Further Details}\label{appendix:details}

\subsection{Results}\label{appendix:results}
We tabulate the complete results for BERT and DistilBERT on ADE, LEDGAR, and PATENT in Tables \ref{table:bert} and \ref{table:distilbert} respectively. The values in each table are averaged across 5 seeds.

\begin{table*}[t]
    \begin{subtable}{\textwidth}
    \centering
        \begin{tabular}{ccccc}
            \hline
            
            \multirow{2}{*}{\textbf{Model}}         & \multicolumn{4}{c}{\textbf{Maximum Sequence Length}}  \\          
            
                                                    & \textbf{128}  & \textbf{64}   & \textbf{32}   & \textbf{16}   \\ 
            
            \hline
    
            $\tokgen$                               & 90.74 $\pm$ 0.84      & 91.22 $\pm$ 0.74      & 87.78 $\pm$ 0.74      & 76.04 $\pm$ 2.09  \\         
            $\tokgen^{1000}$                        & 90.66 $\pm$ 0.70      & 90.62 $\pm$ 0.41      & 88.62 $\pm$ 0.41      & 80.26 $\pm$ 0.91  \\ 
            $\tokgen^{2500}$                        & 91.08 $\pm$ 0.54      & 90.76 $\pm$ 0.87      & 89.06 $\pm$ 0.87      & 80.76 $\pm$ 0.93  \\ 
            $\tokgen^{5000}$                        & 91.00 $\pm$ 0.68      & 91.28 $\pm$ 0.62      & 89.28 $\pm$ 0.62      & 79.92 $\pm$ 1.42  \\ 
            $\tok100$                               & 90.96 $\pm$ 0.67      & 90.82 $\pm$ 0.71      & 89.32 $\pm$ 0.71      & 82.82 $\pm$ 0.85  \\ 
            $\tok100^{1000}$                        & 89.96 $\pm$ 1.21      & 90.38 $\pm$ 0.48      & 89.00 $\pm$ 0.48      & 85.18 $\pm$ 1.11  \\ 
            $\tok100^{2500}$                        & 89.94 $\pm$ 0.72      & 90.56 $\pm$ 0.61      & 89.54 $\pm$ 0.61      & 85.78 $\pm$ 0.72  \\ 
            $\tok100^{5000}$                        & 90.28 $\pm$ 0.65      & 90.38 $\pm$ 0.75      & 90.70 $\pm$ 0.75      & 84.94 $\pm$ 0.45  \\   
            
            \hline
        \end{tabular}
        \caption{ADE}
    \end{subtable}

    \bigskip
    \begin{subtable}{\textwidth}
    \centering
        \begin{tabular}{ccccc}
            \hline
            
            \multirow{2}{*}{\textbf{Model}}         & \multicolumn{4}{c}{\textbf{Maximum Sequence Length}}  \\          
            
                                                    & \textbf{512}  & \textbf{256}   & \textbf{128}   & \textbf{64} \\ 
            
            \hline
    
            $\tokgen$                               & 82.12 $\pm$ 0.33      & 81.94 $\pm$ 0.36      & 81.46 $\pm$ 0.39      & 79.62 $\pm$ 0.56  \\         
            $\tokgen^{1000}$                        & 82.56 $\pm$ 0.24      & 82.52 $\pm$ 0.35      & 82.12 $\pm$ 0.40      & 80.54 $\pm$ 0.37  \\ 
            $\tokgen^{2500}$                        & 82.16 $\pm$ 0.44      & 82.24 $\pm$ 0.40      & 81.92 $\pm$ 0.54      & 80.80 $\pm$ 0.57  \\ 
            $\tokgen^{5000}$                        & 82.08 $\pm$ 0.41      & 82.02 $\pm$ 0.20      & 81.66 $\pm$ 0.19      & 80.70 $\pm$ 0.16  \\ 
            $\tok100$                               & 82.12 $\pm$ 0.41      & 82.34 $\pm$ 0.21      & 81.68 $\pm$ 0.43      & 79.74 $\pm$ 0.66  \\ 
            $\tok100^{1000}$                        & 82.38 $\pm$ 0.58      & 82.30 $\pm$ 0.68      & 81.80 $\pm$ 0.34      & 80.84 $\pm$ 0.23  \\ 
            $\tok100^{2500}$                        & 81.96 $\pm$ 0.57      & 81.78 $\pm$ 0.60      & 82.06 $\pm$ 0.35      & 80.72 $\pm$ 0.57  \\ 
            $\tok100^{5000}$                        & 82.14 $\pm$ 0.58      & 82.32 $\pm$ 0.35      & 81.92 $\pm$ 0.31      & 80.92 $\pm$ 0.71  \\   
            
            \hline
        \end{tabular}
        \caption{LEDGAR}
    \end{subtable}

    \bigskip
    \begin{subtable}{\textwidth}
    \centering
        \begin{tabular}{ccccc}
            \hline
            
            \multirow{2}{*}{\textbf{Model}}         & \multicolumn{4}{c}{\textbf{Maximum Sequence Length}}  \\          
            
                                                    & \textbf{256}  & \textbf{128}   & \textbf{64}   & \textbf{32}  \\ 
            
            \hline
    
            $\tokgen$                               & 61.44 $\pm$ 0.38      & 61.28 $\pm$ 0.37      & 60.46 $\pm$ 0.24      & 58.60 $\pm$ 0.60  \\         
            $\tokgen^{1000}$                        & 61.18 $\pm$ 0.54      & 61.28 $\pm$ 0.36      & 60.40 $\pm$ 0.45      & 59.46 $\pm$ 0.50  \\ 
            $\tokgen^{2500}$                        & 61.40 $\pm$ 0.46      & 61.40 $\pm$ 0.69      & 61.22 $\pm$ 0.68      & 59.26 $\pm$ 0.42  \\ 
            $\tokgen^{5000}$                        & 61.16 $\pm$ 0.69      & 61.08 $\pm$ 0.49      & 60.40 $\pm$ 0.71      & 59.14 $\pm$ 0.44  \\ 
            $\tok100$                               & 60.66 $\pm$ 0.39      & 60.62 $\pm$ 1.04      & 59.52 $\pm$ 0.63      & 58.44 $\pm$ 0.63  \\ 
            $\tok100^{1000}$                        & 60.96 $\pm$ 0.62      & 60.16 $\pm$ 0.68      & 59.48 $\pm$ 0.25      & 58.76 $\pm$ 0.63  \\ 
            $\tok100^{2500}$                        & 60.80 $\pm$ 0.42      & 60.36 $\pm$ 1.02      & 59.98 $\pm$ 1.15      & 58.78 $\pm$ 0.58  \\ 
            $\tok100^{5000}$                        & 60.42 $\pm$ 0.44      & 59.80 $\pm$ 0.73      & 59.54 $\pm$ 0.46      & 58.24 $\pm$ 1.76  \\   
            
            \hline
        \end{tabular}
        \caption{PATENT}
    \end{subtable}

    \caption{Model performance of BERT averaged across 5 seeds.}
    \label{table:bert}
\end{table*}

\begin{table*}[t]
    \begin{subtable}{\textwidth}
    \centering
        \begin{tabular}{ccccc}
            \hline
            
            \multirow{2}{*}{\textbf{Model}}         & \multicolumn{4}{c}{\textbf{Maximum Sequence Length}}  \\          
            
                                                    & \textbf{128}  & \textbf{64}   & \textbf{32}   & \textbf{16}   \\ 
            
            \hline
    
            Distil. + $\tokgen$                     & 90.66 $\pm$ 0.69      & 91.66 $\pm$ 0.43      & 87.56 $\pm$ 1.64      & 74.78 $\pm$ 1.50  \\
            Distil. + $\tokgen^{1000}$              & 90.18 $\pm$ 0.89      & 90.44 $\pm$ 0.73      & 88.16 $\pm$ 0.81      & 78.74 $\pm$ 0.88  \\
            Distil. + $\tokgen^{2500}$              & 91.08 $\pm$ 0.28      & 90.64 $\pm$ 0.53      & 88.30 $\pm$ 0.96      & 79.24 $\pm$ 1.37  \\
            Distil. + $\tokgen^{5000}$              & 89.60 $\pm$ 0.92      & 90.22 $\pm$ 1.11      & 88.06 $\pm$ 0.79      & 79.52 $\pm$ 1.16  \\
            Distil. + $\tok100$                     & 90.52 $\pm$ 0.48      & 89.76 $\pm$ 0.84      & 88.54 $\pm$ 1.01      & 81.16 $\pm$ 0.91  \\
            Distil. + $\tok100^{1000}$              & 88.26 $\pm$ 0.86      & 89.10 $\pm$ 0.44      & 88.52 $\pm$ 0.68      & 82.84 $\pm$ 0.35  \\
            Distil. + $\tok100^{2500}$              & 88.58 $\pm$ 1.20      & 89.10 $\pm$ 1.18      & 89.32 $\pm$ 1.01      & 83.38 $\pm$ 0.62  \\
            Distil. + $\tok100^{5000}$              & 87.68 $\pm$ 0.92      & 87.94 $\pm$ 1.22      & 87.88 $\pm$ 0.55      & 82.84 $\pm$ 0.77  \\
            
            \hline
        \end{tabular}
        \caption{ADE}
    \end{subtable}

    \bigskip
    \begin{subtable}{\textwidth}
    \centering
        \begin{tabular}{ccccc}
            \hline
            
            \multirow{2}{*}{\textbf{Model}}         & \multicolumn{4}{c}{\textbf{Maximum Sequence Length}}  \\          
            
                                                    & \textbf{512}  & \textbf{256}   & \textbf{128}   & \textbf{64} \\ 
            
            \hline
    
            Distil. + $\tokgen$                     & 81.48 $\pm$ 0.52      & 81.12 $\pm$ 0.50      & 81.18 $\pm$ 0.31      & 79.22 $\pm$ 0.29  \\
            Distil. + $\tokgen^{1000}$              & 82.02 $\pm$ 0.83      & 82.30 $\pm$ 0.31      & 81.56 $\pm$ 0.44      & 80.20 $\pm$ 0.41  \\
            Distil. + $\tokgen^{2500}$              & 81.74 $\pm$ 0.23      & 81.36 $\pm$ 0.25      & 81.86 $\pm$ 0.18      & 79.90 $\pm$ 1.01  \\
            Distil. + $\tokgen^{5000}$              & 81.38 $\pm$ 0.52      & 81.62 $\pm$ 0.29      & 81.60 $\pm$ 0.29      & 80.34 $\pm$ 0.28  \\
            Distil. + $\tok100$                     & 81.42 $\pm$ 0.70      & 81.60 $\pm$ 0.12      & 81.50 $\pm$ 0.48      & 80.02 $\pm$ 0.54  \\
            Distil. + $\tok100^{1000}$              & 81.42 $\pm$ 0.59      & 80.90 $\pm$ 0.68      & 81.98 $\pm$ 0.18      & 80.62 $\pm$ 0.47  \\
            Distil. + $\tok100^{2500}$              & 81.80 $\pm$ 0.17      & 81.36 $\pm$ 0.30      & 82.06 $\pm$ 0.27      & 80.46 $\pm$ 0.38  \\
            Distil. + $\tok100^{5000}$              & 81.58 $\pm$ 0.57      & 81.34 $\pm$ 0.42      & 81.92 $\pm$ 0.18      & 80.82 $\pm$ 0.43  \\
            
            \hline
        \end{tabular}
        \caption{LEDGAR}
    \end{subtable}

    \bigskip
    \begin{subtable}{\textwidth}
    \centering
        \begin{tabular}{ccccc}
            \hline
            
            \multirow{2}{*}{\textbf{Model}}         & \multicolumn{4}{c}{\textbf{Maximum Sequence Length}}  \\          
            
                                                    & \textbf{256}  & \textbf{128}   & \textbf{64}   & \textbf{32}  \\ 
            
            \hline
    
            Distil. + $\tokgen$                     & 60.88 $\pm$ 0.61      & 60.98 $\pm$ 0.67      & 59.88 $\pm$ 0.57      & 57.72 $\pm$ 0.71  \\
            Distil. + $\tokgen^{1000}$              & 60.58 $\pm$ 0.31      & 59.92 $\pm$ 0.63      & 59.94 $\pm$ 0.94      & 58.36 $\pm$ 0.62  \\
            Distil. + $\tokgen^{2500}$              & 59.96 $\pm$ 0.75      & 59.94 $\pm$ 0.43      & 59.90 $\pm$ 0.65      & 58.16 $\pm$ 0.61  \\
            Distil. + $\tokgen^{5000}$              & 59.86 $\pm$ 0.61      & 60.10 $\pm$ 0.88      & 59.26 $\pm$ 0.53      & 58.46 $\pm$ 0.52  \\
            Distil. + $\tok100$                     & 59.58 $\pm$ 0.77      & 59.22 $\pm$ 0.59      & 58.10 $\pm$ 0.70      & 57.22 $\pm$ 0.59  \\
            Distil. + $\tok100^{1000}$              & 59.52 $\pm$ 0.49      & 59.88 $\pm$ 0.54      & 58.72 $\pm$ 0.47      & 57.42 $\pm$ 0.72  \\
            Distil. + $\tok100^{2500}$              & 59.04 $\pm$ 0.32      & 58.82 $\pm$ 0.95      & 57.58 $\pm$ 0.53      & 56.76 $\pm$ 0.47  \\
            Distil. + $\tok100^{5000}$              & 59.82 $\pm$ 0.57      & 58.74 $\pm$ 0.40      & 58.76 $\pm$ 0.59      & 57.30 $\pm$ 1.01  \\
            
            \hline
        \end{tabular}
        \caption{PATENT}
    \end{subtable}

    \caption{Model performance of DistilBERT averaged across 5 seeds.}
    \label{table:distilbert}
\end{table*}

% This is a section in the appendix.

\end{document}